\documentclass{article}

\usepackage{PRIMEarxiv}

\usepackage[utf8]{inputenc} 
\usepackage[T1]{fontenc}    
\usepackage{hyperref}       
\usepackage{booktabs}       
\usepackage{amsfonts}       
\usepackage{nicefrac}       
\usepackage{microtype}      
\usepackage{lipsum}
\usepackage{fancyhdr}       
\usepackage{graphicx}       
\usepackage{times}  
\usepackage{helvet} 
\usepackage{courier}  

\usepackage{graphicx} 
\usepackage{natbib}  
\usepackage{caption} 
\usepackage{xcolor}
\usepackage{textcomp}
\usepackage{subfig}
\graphicspath{{media/}}     

\title{TAILOR: Teaching with Active and Incremental Learning for Object Registration
}

\author{
  Qianli Xu, 
    Nicolas Gauthier, 
    Wenyu Liang, 
    Fen Fang, 
    Hui Li Tan, 
    Ying Sun, \\
    Yan Wu, 
    Liyuan Li, 
    Joo-Hwee Lim \\ 
  Institute for Infocomm Research, 
    Agency for Science, Technology and Research, Singapore \\
  \texttt{Email: qxu@i2r.a-star.edu.sg} \\
}

\begin{document}
\maketitle

\begin{abstract}

When deploying a robot to a new task, one often has to train it to detect novel objects, which is  time-consuming and labor-intensive. We present TAILOR - a method and system for object registration with active and incremental learning. When instructed by a human teacher to register an object, TAILOR is able to automatically select viewpoints to capture informative images by actively exploring viewpoints, and employs a fast incremental learning algorithm to learn new objects without potential forgetting of previously learned objects. We demonstrate the effectiveness of our method with a KUKA robot to learn novel objects used in a real-world gearbox assembly task through natural interactions.
\end{abstract}

\section{Introduction}
In many industrial and domestic applications, automating object detection and recognition is an essential capability. Fast object instance learning becomes a fundamental problem and key requirement of automated systems and machines. Recently, deep learning-based methods have achieved impressive detection accuracy. However, when deployed for a new task, the system needs to be re-trained on new objects, which requires costly human efforts to collect and annotate a large number of images. 
Some works deal with the generation of high-quality training data by (i) simplifying the data collection and annotation process using specific devices or user interfaces~\cite{Marion2018LabelFA}, or (ii) automating the generation of data samples through augmentation and synthesis~\cite{Dwibedi2017CutPA}. For the former, it still involves a lengthy process of data sampling and manual annotation. For the latter, the characteristics of synthesized data may not match with those of actual data from domain-specific environments, leading to unstable and deteriorated performance.

\begin{figure}[t]
\centering
\includegraphics[width=0.69\linewidth]{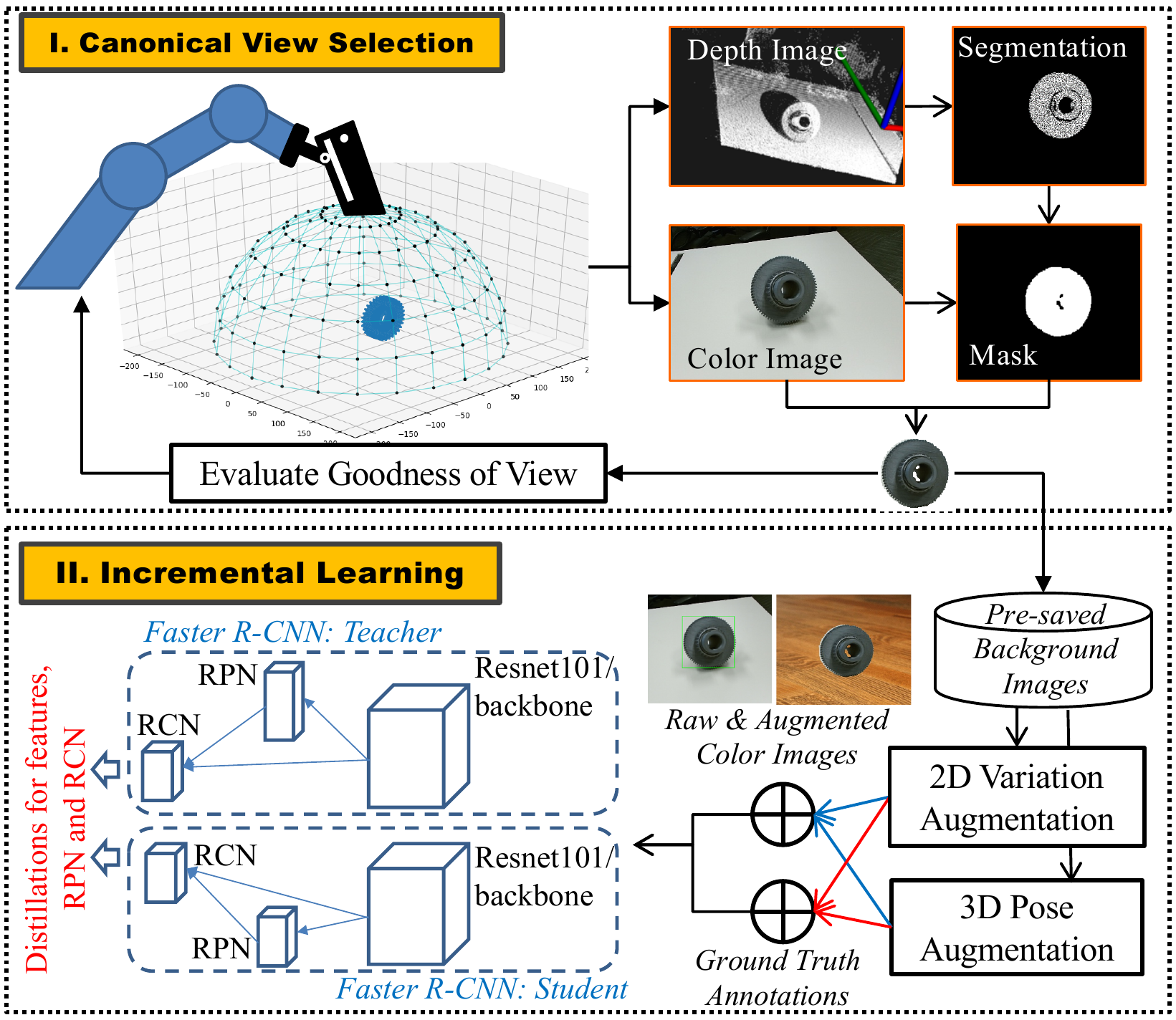} 
\caption{A system architecture of TAILOR}
\label{arch}
\end{figure}

\begin{figure*}[t]
\centering
\includegraphics[width=0.99\linewidth]{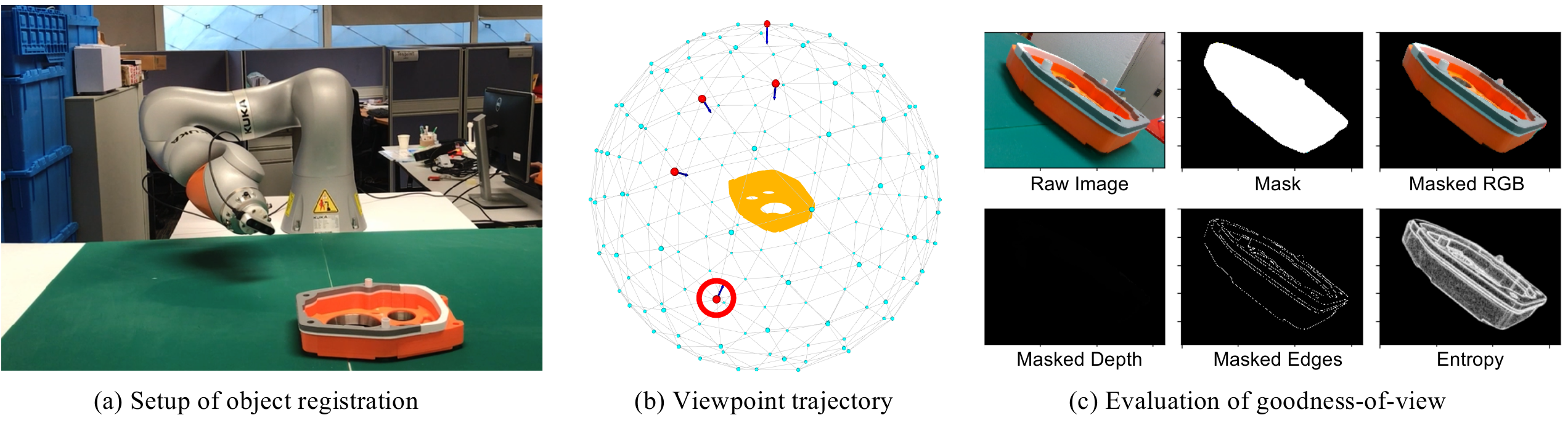} 
\caption{Object registration with canonical view selection}
\label{fig:registration}
\end{figure*}

\textbf{Related work} To simplify data collection and annotation, a few works resort to interactive data collection and annotation, where an agent is endowed with the ability to register object instances with human guidance~\cite{Kasaei2015InteractiveOL,Dehghan2019OnlineOA,Kasaei2018PerceivingLA}. However, they usually require strong prior knowledge of objects (e.g., 3D models) and are restricted by the limited functionalities of the hardware (\emph{e.g.}, robot mobility) and software, thus only addressing small-scale toy problems. It is paramount to fill this gap with an effective interaction protocol for agents to learn new objects in a similar way as a human learner does.
Rooted in object detection, object learning can be boosted by using preferential viewpoint information~\cite{Xu_icip20}. For example, the speed and accuracy of object recognition are higher in canonical views relative to non-canonical views, as shown in experiments with both computers and humans~\cite{Ghose2013GeneralizationBC,Poggio1990ANT}. Here, canonical views refer to viewpoints that visually characterize the entity of interest, which often encompass stable, salient and significant visual features~\cite{Blanz1999WhatOA}. 

This paper demonstrates an interactive process facilitated by TAILOR - a method and system that enables \textit{Teaching with Active and Incremental Learning for Object Registration}. Using TAILOR, a robot learns new object instances while leveraging the representation bias of canonical views. 
We show how a robot with an eye-in-hand camera interacts with a human teacher to register new objects through (1) human instruction grounding, (2) active image sampling based on canonical view selection, and (3) fast training of an object detector with incremental learning. We present the learning outcome in a real-world use case of interactive object detection. Our method can greatly shorten the cycle-time and reduce manpower cost in new object registration, thus contributing to fast and low-cost deployment of vision-based solutions in real-world scenarios.

\section{Method}
The system architecture of TAILOR is shown in Figure~\ref{arch}. It consists of two phases. In phase I - canonical view selection - the robot uses an eye-in-hand RealSense D435 camera to get RGB-D images from selected viewpoints, where the viewpoints are evaluated for their goodness of view (GOV) in real-time. The GOV is used to determine canonical views where the RGB images are registered as training samples. In phase II, the selected samples are fed into the training pipeline with a series of data augmentation mechanisms and an incremental learning scheme.

\subsection{Interaction Protocol} 
To register new objects, a human teacher issues speech commands to activate the teaching process. For example, the teacher puts an object on the tabletop and says ``\textit{Start object registration.''} Then, the teacher points at the object and says ``\textit{This is the input shaft.}". Through keyword matching and object segmentation, the system registers the object name and relates it with the new incoming training samples. Figure~\ref{fig:registration}(a) shows the setup of object registration. To test the object detector, the teacher may put one or a few objects on the table and ask ``\textit{Where is the input shaft?}''. The robot performs detection to locate the corresponding object and displays visual answer (bounding box) or points at it with the gripper. 

\subsection{Canonical View Selection}
\label{sec:cvs}
\begin{itemize}
    \item{\textbf{Segmentation}} The RealSense D435 camera provides a color image and point cloud of the scene. Using the point cloud, we adopt the Point Cloud Library (PCL) to estimate the dominant plane model and extract the object on the tabletop. This results in a regional mask of the object, which is used to extract both the color information and depth information within the object region.
    \item{\textbf{Goodness of view}}
    The \emph{goodness} of a viewpoint can be measured by multiple metrics defined on different visual features, such as, visible area, silhouette, depth, visual stability, curvature entropy, mesh saliency, \emph{etc.}~\cite{Polonsky2005WhatS,Dutagaci2010ABF}. The GOV is a function of the statistical distributions of visual features that denote its informativeness. To compute GOV, we first perform aforementioned depth segmentation. Next, we use the resultant mask to apply on the color and depth images to extract the regional data of the object. Finally, we compute the GOV based on multiple visual features, including silhouette length, depth distribution, curvature entropy, and color entropy~\cite{Xu_icip20}. Canonical views are registered as those with higher combined GOV.
    \item{\textbf{Viewpoint exploration}}
    Unlike other interactive learning methods that rely on prior 3D information of an object, our canonical view selection method gathers sparse viewpoint information and evaluates the aggregated GOV of the viewpoints on-the-fly. This denotes a learning experience similar to a robot exploring an unknown object or space. 
    The only assumption is that an RGB-D camera is calibrated with respect to a set of candidate viewpoints on a virtual spherical surface, where an object of interest is located at the sphere center. Thus, the 3D coordinates of the viewpoints are known and the robot can move the camera to those points. The robot then follows the OnLIne Viewpoint Exploration (OLIVE) routine proposed in ~\cite{Xu_icip20} to visit viewpoints. Figure~\ref{fig:registration}(b) shows the view trajectory for an industry object. The RGB-D data captured at each viewpoint is used to extract object features and compute the GOV. Figure~\ref{fig:registration}(c) shows the visual features for GOV evaluation at a particular viewpoint. At any viewpoint, the robot searches the local maxima of GOV, where the GOV is computed as the weighted sum of individual GOV metrics. Once the local maxima is found, the next view is chosen as one with the largest geographical distance to the current view. 
\end{itemize}

In the current setup, we put a single object on the table. However, this can be extended to multiple objects with two additional components: (1) multiple object segmentation, and (2) hand gesture recognition. For (1), all objects are assumed to be placed in separation and a unique object mask can be extracted for individual objects. With (2) we can extract the pointing gesture so as get the object of interest, i.e., the one being pointed at.

\subsection{Incremental Learning}
\label{sec:inclrn}

The image samples from canonical views are augmented through 2D variation and 3D transformation to generate a training set ~\cite{FangF2019SelfTeach}. 
Once the training images of one or a few novel objects have been generated, the system performs on-site incremental learning by calling an incremental learning algorithm on a state-of-the-art generic object detector, such as Fast R-CNN~\cite{fastrcnn} and Faster R-CNN ~\cite{fasterrcnn}. 
As examples, the incremental learning algorithm by~\cite{shmelkov2017} employs biased distillation for Fast R-CNN while the recently developed fast incremental learning algorithm for Faster R-CNN (Faster ILOD)~\cite{Peng2020} uses knowledge distillation for object detectors based on Region Proposal Networks (RPNs) to achieve efficient end-to-end learning. These incremental learning algorithms employ distillation approaches to avoid catastrophic forgetting. There would be small reduction in accuracy on the newly learned object compared to normal re-training. The incremental learning algorithm is able to avoid catastrophic forgetting of previously learned objects when re-trained just for new objects. The updated model is then applied to detect new objects in new tasks, while still capable of detecting objects learned previously.

\section{Evaluation}
We implement TAILOR with a KUKA LBR iiwa 14 R820 robot of 7-degree-of-freedom (7-DoF) and a RealSense D435 camera mounted on the end-effector. The KUKA robot is a lightweight robot which offers high flexible motion and high-performance servo control with repeatability of $\pm 0.15$ mm. We pre-define a set of 88 viewpoints at the vertices of an icosahedron on a semi-sphere (radius = 350 mm). The geometric center of the object is aligned with the viewpoint sphere center. The camera position is calibrated with a default top view position and its pose is preset so as to point towards the object center at any viewpoint. 
For a flat object, it is flipped during the registration process so that images of the other side are captured. We perform interactive learning on a few novel components used for the assembly of an industrial gearbox product.

\begin{figure}[t]
\centering
\includegraphics[width=0.6\columnwidth]{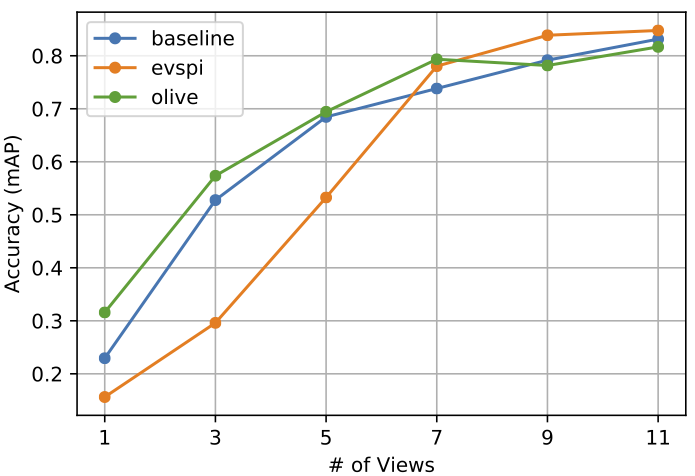} 
\caption{Comparing viewpoint selection on T-LESS dataset}
\label{fig:benchmark}
\end{figure}

For benchmarking, we evaluate the effect of canonical view selection on the T-LESS dataset~\cite{Hodan2017TLESSAR}. We compare our method against a baseline random viewpoint selection method and an efficient view selection using proximity information (EVSPI)~\cite{Gao2016EfficientVS}. The detection performance is shown in Figure.~\ref{fig:benchmark}. Using only a few image samples, our method is able to train object detectors that outperform alternative methods, especially when the view budget is low. Interested readers may refer to~\cite{Xu_icip20} for detailed benchmarking results.

\begin{figure}[t]
\centering
\subfloat[Human view]{\includegraphics[width=0.51\columnwidth]{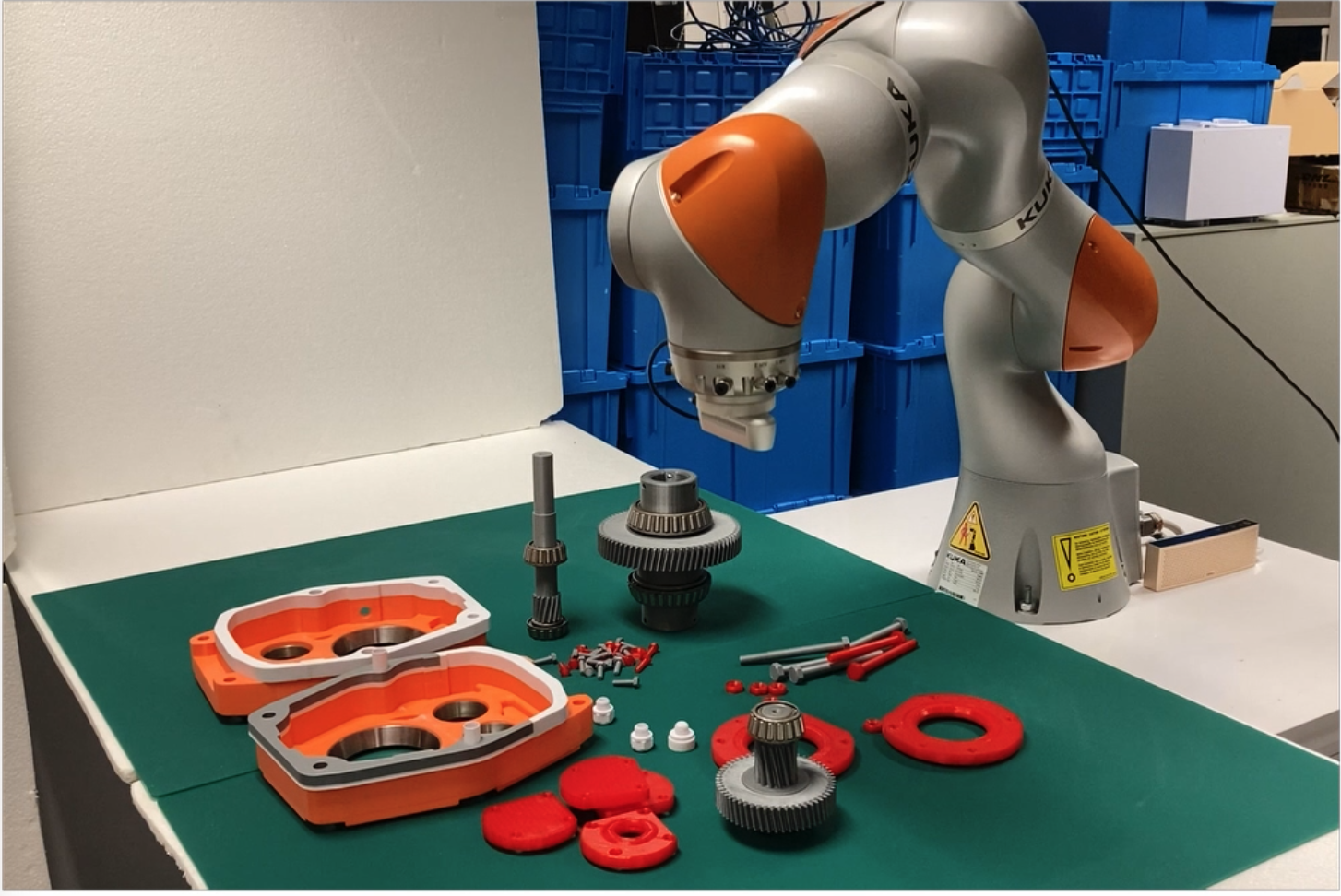}}
\subfloat[Robot view]{\includegraphics[width=0.455\columnwidth]{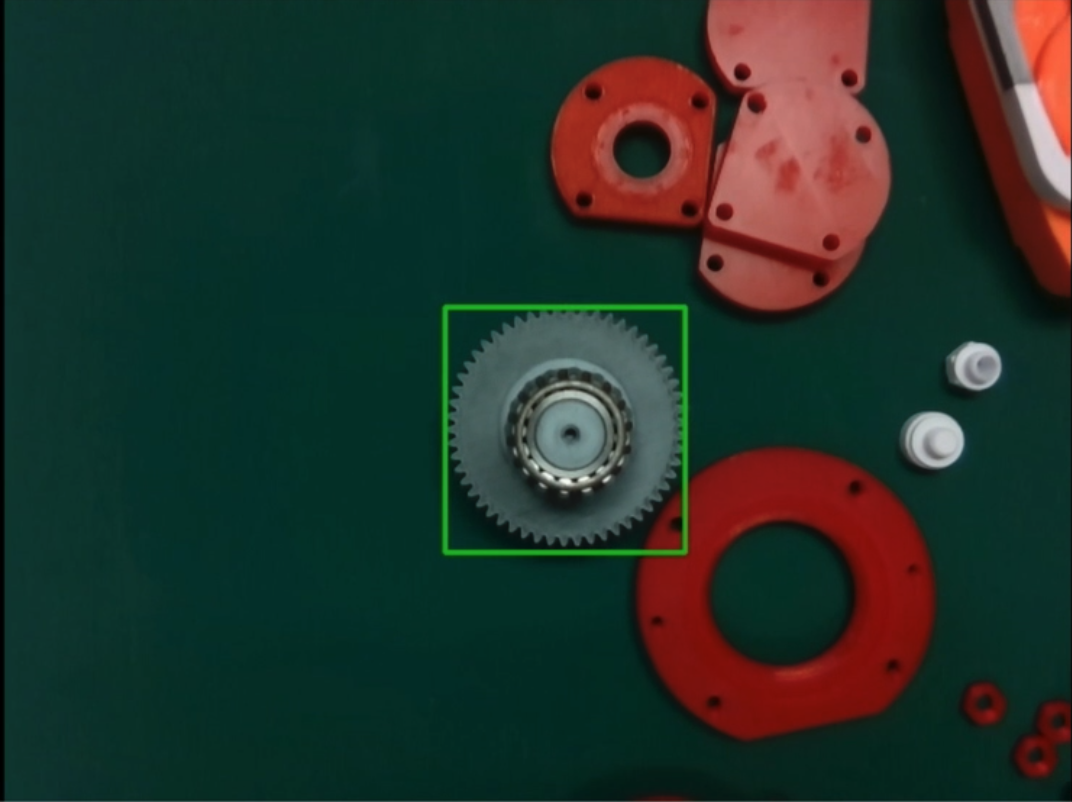}}\\

\caption{Robot detecting object ``\emph{transfer shaft}''}
\label{fig:detect}
\end{figure}

For testing, we put multiple objects (including unknown objects) on the tabletop. The robot is able to reliably detect learned objects and provide corresponding feedback through speech, visual display and pointing gestures. Figure~\ref{fig:detect} is a snapshot of the robot detecting an object ``\emph{transfer shaft}''.

\section{Conclusions and Future Work}
TAILOR provides an intuitive and efficient way to learn new objects for autonomous agents. It can be used by non-expert users to teach robots new objects without efforts of programming or annotation. As such, the method provides a scalable solution to real-world tasks involving vision-based object detection. To enhance scalability, we intend to simplify the setup by (1) alleviating the need of a dominant flat surface to support the object (which is required for depth-based object segmentation), (2) adopting a mobile robot to allow for more flexible and larger ranges of viewpoint exploration, and (3) employing more powerful natural speech recognition and synthesis (vs. command-type keyword matching) to enhance the human-robot interaction.

\section{Acknowledgments}
This research is supported by the Agency for Science, Technology and Research (A*STAR) under AME Programmatic Funding Scheme (Project No. A18A2b0046).
\bibliographystyle{unsrt}  
\bibliography{references} 

\begin{thebibliography}{18}
\providecommand{\natexlab}[1]{#1}
\providecommand{\url}[1]{\texttt{#1}}
\providecommand{\urlprefix}{URL }
\expandafter\ifx\csname urlstyle\endcsname\relax
  \providecommand{\doi}[1]{doi:\discretionary{}{}{}#1}\else
  \providecommand{\doi}{doi:\discretionary{}{}{}\begingroup
  \urlstyle{rm}\Url}\fi

\bibitem[{Blanz, Tar, and B{\"u}lthoff(1999)}]{Blanz1999WhatOA}
Blanz, V.; Tar, M.; and B{\"u}lthoff, H. 1999.
\newblock What object attributes determine canonical views?
\newblock \emph{Perception} 28 5: 575--99.

\bibitem[{Dehghan et~al.(2019)Dehghan, Zhang, Siam, Jin, Petrich, and
  J{\"a}gersand}]{Dehghan2019OnlineOA}
Dehghan, M.; Zhang, Z.; Siam, M.; Jin, J.; Petrich, L.; and J{\"a}gersand, M.
  2019.
\newblock Online object and task learning via human robot interaction.
\newblock \emph{2019 International Conference on Robotics and Automation
  (ICRA)} 2132--2138.

\bibitem[{Dutagaci, Cheung, and Godil(2010)}]{Dutagaci2010ABF}
Dutagaci, H.; Cheung, C.; and Godil, A. 2010.
\newblock A benchmark for best view selection of 3D objects.
\newblock In \emph{3DOR '10}.

\bibitem[{Dwibedi, Misra, and Hebert(2017)}]{Dwibedi2017CutPA}
Dwibedi, D.; Misra, I.; and Hebert, M. 2017.
\newblock Cut, paste and learn: Surprisingly easy synthesis for instance
  detection.
\newblock \emph{2017 IEEE International Conference on Computer Vision (ICCV)}
  1310--1319.

\bibitem[{Fang et~al.(2019)Fang, Xu, Cheng, Li, Sun, and
  Lim}]{FangF2019SelfTeach}
Fang, F.; Xu, Q.; Cheng, Y.; Li, L.; Sun, Y.; and Lim, J.-H. 2019.
\newblock Self-teaching strategy for learning to recognize novel objects in
  collaborative robots.
\newblock In \emph{ICRAI'19}.

\bibitem[{Gao, Wang, and Han(2016)}]{Gao2016EfficientVS}
Gao, T.; Wang, W.; and Han, H. 2016.
\newblock Efficient view selection by measuring proxy information.
\newblock \emph{Journal of Visualization and Computer Animation} 27: 351--357.

\bibitem[{Ghose and Liu(2013)}]{Ghose2013GeneralizationBC}
Ghose, T.; and Liu, Z. 2013.
\newblock Generalization between canonical and non-canonical views in object
  recognition.
\newblock \emph{Journal of Vision} 13 1.

\bibitem[{Girshick(2015)}]{fastrcnn}
Girshick, R. 2015.
\newblock Fast R-CNN.
\newblock In \emph{International Conference on Computer Vision ({ICCV})}.

\bibitem[{Hodan et~al.(2017)Hodan, Haluza, Obdrz{\'a}lek, Matas, Lourakis, and
  Zabulis}]{Hodan2017TLESSAR}
Hodan, T.; Haluza, P.; Obdrz{\'a}lek, S.; Matas, J. E.~S.; Lourakis, M. I.~A.;
  and Zabulis, X. 2017.
\newblock T-LESS: An RGB-D Dataset for 6D Pose Estimation of Texture-Less
  Objects.
\newblock \emph{2017 IEEE Winter Conference on Applications of Computer Vision
  (WACV)} 880--888.

\bibitem[{Kasaei et~al.(2015)Kasaei, Oliveira, Lim, Lopes, and
  Tom{\'e}}]{Kasaei2015InteractiveOL}
Kasaei, S.; Oliveira, M.; Lim, G.; Lopes, L.; and Tom{\'e}, A. 2015.
\newblock Interactive open-ended learning for 3d object recognition: An
  approach and experiments.
\newblock \emph{Journal of Intelligent \& Robotic Systems} 80: 537--553.

\bibitem[{Kasaei et~al.(2018)Kasaei, Sock, Lopes, Tom{\'e}, and
  Kim}]{Kasaei2018PerceivingLA}
Kasaei, S.; Sock, J.; Lopes, L.; Tom{\'e}, A.; and Kim, T.-K. 2018.
\newblock Perceiving, learning, and recognizing 3d objects: An approach to
  cognitive service robots.
\newblock In \emph{AAAI}.

\bibitem[{Marion et~al.(2018)Marion, Florence, Manuelli, and
  Tedrake}]{Marion2018LabelFA}
Marion, P.; Florence, P.; Manuelli, L.; and Tedrake, R. 2018.
\newblock Label fusion: A pipeline for generating ground truth labels for real
  rgbd data of cluttered scenes.
\newblock \emph{2018 IEEE International Conference on Robotics and Automation
  (ICRA)} 1--8.

\bibitem[{Peng, Zhao, and Lovell(2020)}]{Peng2020}
Peng, C.; Zhao, K.; and Lovell, B. 2020.
\newblock Faster ILOD: Incremental Learning for Object Detectors based on
  Faster RCNN.
\newblock \emph{ArXiv} abs/2003.03901.

\bibitem[{Poggio and Edelman(1990)}]{Poggio1990ANT}
Poggio, T.; and Edelman, S. 1990.
\newblock A network that learns to recognize three-dimensional objects.
\newblock \emph{Nature} 343: 263--266.

\bibitem[{Polonsky et~al.(2005)Polonsky, Patan{\`e}, Biasotti, and
  Gotsman}]{Polonsky2005WhatS}
Polonsky, O.; Patan{\`e}, G.; Biasotti, S.; and Gotsman, C. 2005.
\newblock What's in an image? Towards the computation of the ``best'' view of
  an object.

\bibitem[{Ren et~al.(2015)Ren, He, Girshick, and Sun}]{fasterrcnn}
Ren, S.; He, K.; Girshick, R.; and Sun, J. 2015.
\newblock Faster {R-CNN}: Towards Real-Time Object Detection with Region
  Proposal Networks.
\newblock In \emph{Advances in Neural Information Processing Systems ({NIPS})}.

\bibitem[{Shmelkov, Schmid, and Alahari(2017)}]{shmelkov2017}
Shmelkov, K.; Schmid, C.; and Alahari, K. 2017.
\newblock Incremental Learning of Object Detectors without Catastrophic
  Forgetting.
\newblock \emph{CoRR} abs/1708.06977.
\newblock \urlprefix\url{http://arxiv.org/abs/1708.06977}.

\bibitem[{Xu et~al.(2020)Xu, Fang, Gauthier, Li, and Lim}]{Xu_icip20}
Xu, Q.; Fang, F.; Gauthier, N.; Li, L.; and Lim, J.-H. 2020.
\newblock Active image sampling on canonical views for novel object detection.
\newblock In \emph{ICIP'20}.

\end{thebibliography}

\end{document}